\documentclass[final]{nesy2026}

\usepackage{booktabs}
\usepackage{pifont}

\newcommand{\bench}{ClosureBench}
\newcommand{\owa}{OWA}
\newcommand{\cwa}{CWA}
\newcommand{\lcwa}{LCWA}
\newcommand{\closure}{\operatorname{Cl}}
\newcommand{\cmark}{\ding{51}}
\newcommand{\xmark}{\ding{55}}
\newcommand{\pmark}{$\sim$}
\usepackage{colortbl}

\title[\bench]{\bench:\\ A Benchmark for Reasoning under Closure Assumptions}

\author{%
  \Name{Francesco Belardinelli}
  \Email{francesco.belardinelli@imperial.ac.uk}\\
  \addr Imperial College London
  \AND
  \Name{Flavio Corradini}
  \Email{flavio.corradini@unicam.it}\\
  \Name{Matteo Leonesi\nametag{\thanks{Corresponding author.}}}
  \Email{matteo.leonesi@unicam.it}\\
  \addr University of Camerino
}
\begin{document}

\maketitle

\begin{abstract}
AI agents built on large language models (LLMs) often answer questions from incomplete records. Missing information might be interpreted as \texttt{unknown} or \texttt{false}, depending on the completeness of the source. Current evaluations fail to distinguish these two aspects, as they are usually interwoven with other reasoning demands. We present \bench{}, a benchmark that tests whether LLMs follow the stated rules for dealing with missing information. Every base scenario is tested under \owa{}, \cwa{}, and \lcwa{}, and only the treatment of missing information changes between variants. \bench{} provides a verified base benchmark as well as agent extensions for action choice, source scope, and updates. A model is considered correct on a base scenario only if it provides correct answers in all of its closure variants. Across 960 instances, overall accuracy can mask substantial closure mistakes. Models achieve overall accuracy from $71.18\%$ to $99.86\%$ across replicated runs, but their core switch accuracy, measured on the scenarios whose gold label actually changes with the contract, varies between $18.23\%$ and $99.31\%$. Large discrepancies emerge under \lcwa{}. For example, some agents choose \texttt{unknown} despite the prompt indicating that missing information should be treated as \texttt{false}. Our findings demonstrate that evaluations of AI agents should not only measure whether agents can reason over existing facts but also whether they use the appropriate assumption when information is absent.
\end{abstract}

\section{Introduction}
\label{sec:intro}
AI agents often look up incomplete external sources before acting \citep{xu2025theagentcompany,miserendino2025swelancer,barres2025tau2bench,qi2025agentif,vijayvargiya2026openagentsafety}. Missing information can be read either as unknown or as false. Consider a deployment request where the record says that a service passed unit tests, but says nothing about whether the service has deployment approval. If the record is incomplete, the approval status remains unknown. If the approval predicate is complete, the missing approval counts as false, even though the visible facts are unchanged. Only the rule for reading the missing atom changes. We call this rule the \emph{closure contract}. It specifies how to interpret missing atoms under the Open World Assumption (\owa{}), the Closed World Assumption (\cwa{}), or the Local Closed World Assumption (\lcwa{}). Closure assumptions have long been studied in the symbolic AI tradition, especially in knowledge representation and database theory \citep{reiter1978closed,smith2004owl,darari2014rdf,razniewski2023completeness,oudshoorn2026shacl}. The same issue becomes practical for LLM-based agents that operate over partial records, where symbolic closure assumptions determine whether the absence of a fact should remain unresolved, support a negative answer, or justify an action. Under \owa{}, an atom that is not stated or derivable remains \texttt{unknown}. Under \cwa{}, such an atom is taken to be \texttt{false}. Under \lcwa{}, this negative inference is allowed only for predicates declared complete. Prior work characterizes when a knowledge base is complete enough to license these negative conclusions \citep{darari2014rdf,razniewski2023completeness}, and recent work on ontology validation shows that the gap between \owa{} reasoning and \cwa{} validation remains important in data systems \citep{oudshoorn2026shacl}. Existing reasoning and agent benchmarks often combine this issue with other abilities \citep{zhu2025bestpractices,kirichenko2025abstentionbench,dou2026deonticbench,hasan2026modalbench}. As a result, it is difficult to tell whether a model fails because it cannot apply the rules or because it does not respect the intended closure policy for unstated and underivable atoms.

\paragraph{Contribution.} We introduce \bench{}, a benchmark for testing whether models follow explicit closure contracts. Each instance defines a small rule base centered on a target atom, together with the predicate vocabulary and the predicates listed as complete. The prompt states the active closure contract, so the model does not need to infer which predicates are complete. For each base scenario, we generate \owa{}, \cwa{}, and \lcwa{} variants. The symbolic case stays fixed, and only the interpretation of unstated and underivable atoms changes.
\begin{itemize}
  \item We introduce \bench, a benchmark for testing \owa{}, \cwa{}, and \lcwa{} reasoning. Each label is verified from the symbolic specification and closure contract. The benchmark also includes three agent extensions. \emph{Ask/Act} turns truth labels into actions, \emph{SourceScope} moves the task to a multi-agent source setting, and \emph{ClosureShift} moves it to a multi-turn update setting. These settings matter because agents rarely answer truth queries in isolation. The extensions test whether a closure error carries over into the decisions an agent makes.
  
  \item We introduce closure switch accuracy, which counts a scenario as correct only when the model answers all of its variants correctly. Grouping instances this way follows behavioral testing \citep{ribeiro2020checklist} and shows that high accuracy can still hide closure errors.
  
  \item We use \bench{} to characterize when closure reasoning fails. The results show that \lcwa{} is the hardest setting because the model must close only the predicates declared complete. In the agent extensions, the same errors affect decisions. In Ask/Act, models ask for more information when they should deny. In SourceScope, they sometimes ignore that the complete source already supports a false answer, or treat absence from an incomplete source as false. In ClosureShift, they fail to keep valid local closure after updates.
\end{itemize}
Code and data are available on \href{https://github.com/MatteoLeonesi/ClosureBench}{GitHub}.
%Data and code availability are described in \appendixref{apd:repro}.

\section{Related work}
\label{sec:related}

%This section summarizes the research areas most closely related to \bench{} and shows how it differs from existing work.

\paragraph{Open world, closed world, and locally closed world assumptions.}
Recent knowledge representation (KR) work studies how open and closed predicates affect validation and query answering under incomplete data \citep{oudshoorn2026shacl,ahmetaj2024cqa,marconi2025openclosed}. \bench{} uses this distinction as an explicit closure contract in the prompt. The facts stay the same, while the closure contract changes.

\paragraph{Logic and rule reasoning benchmarks.}
Recent logic benchmarks evaluate controlled logic and rule reasoning tasks \citep{han2024folio,patel2024multilogieval,wang2024reasonrules,wang2024symbolicwm,zhou2025rulearena,wei2025satbench,chen2025justlogic,qi2025provergen,dou2026deonticbench,hasan2026modalbench,rabern2026logicskills}. Most of these works test reasoning under one fixed semantics. \bench{} keeps the symbolic case fixed and tests whether the model changes its answer only when the closure contract changes.

\paragraph{Abstention, ambiguity, and uncertainty.}
Abstention and clarification work studies cases where the available evidence is not enough for a direct answer \citep{liu2024unknownbench,zhang2024clamber,zhang2025clarify,luo2025clarifymtbench,su2026knowing,liu2025confusebench,kirichenko2025abstentionbench}. \bench{} treats \texttt{unknown} more strictly. It is a label implied by the stated closure contract, not a general uncertainty score.

\paragraph{Agent benchmarks and source scope.}
Recent agent benchmarks test broad agent capabilities in operational settings \citep{yao2024taubench,barres2025tau2bench,lu2025toolsandbox,debenedetti2024agentdojo,diao2025guidebench,qi2025agentif,xu2025theagentcompany,miserendino2025swelancer,vijayvargiya2026openagentsafety,zhu2025bestpractices,liu2023agentbench,zhou2023webarena}. \bench{} complements these suites by isolating source scope. Absence from a complete source can support denial, while absence from an incomplete source should stay unknown. \tableref{tab:positioning} compares the closest benchmarks along the features that separate \bench{} from them and shows that \bench{} is the only suite that keeps the symbolic case fixed while changing only the closure contract, then carries that distinction into \lcwa{} scope and the agent extensions.

\begin{table}[t]
\floatconts
  {tab:positioning}
  {\caption{Comparison with related benchmarks. Gold labels mean the benchmark supports automatic scoring. Insufficient evidence covers expected outputs such as \texttt{unknown}, clarification questions, or requests for information. Scenario switch means that the closure contract changes. Agent extensions cover Ask/Act, SourceScope, and ClosureShift, defined in \sectionref{sec:extensions}. \cmark{} marks a central target, \pmark{} a partial one, and \xmark{} a feature not targeted.}}
  {\small
  \setlength{\tabcolsep}{4pt}
  \begin{tabular*}{\linewidth}{@{\extracolsep{\fill}}lccccc@{}}
  \toprule
  Work & Gold & \shortstack{Insufficient\\evidence} & \shortstack{Scenario\\switch} & \shortstack{LCWA\\scope} & \shortstack{Agent\\extensions} \\
  \midrule
  FOLIO \citep{han2024folio} & \cmark & \cmark & \xmark & \xmark & \xmark \\
  DeonticBench \citep{dou2026deonticbench} & \cmark & \xmark & \xmark & \xmark & \xmark \\
  ModalBench \citep{hasan2026modalbench} & \cmark & \xmark & \xmark & \xmark & \xmark \\
  CLAMBER \citep{zhang2024clamber} & \cmark & \cmark & \xmark & \xmark & \pmark \\
  $\tau$-bench \citep{yao2024taubench} & \cmark & \pmark & \xmark & \xmark & \pmark \\
  \bench{} & \cmark & \cmark & \cmark & \cmark & \cmark \\
  \bottomrule
  \end{tabular*}}
\end{table}

\section{The \bench\ Method}
\label{sec:method}

We first define the closure semantics that fix the gold labels
(\sectionref{sec:semantics}), then describe the dataset (\sectionref{sec:benchmark}), and finally present three agent extensions (\sectionref{sec:extensions}).

\subsection{Closure Semantics}
\label{sec:semantics}

We use \emph{instance} for a single prompt with its gold label and \emph{base
scenario} for the shared symbolic case. Each base scenario has three
instances, one per contract. Each instance contains positive ground atoms $F^+$, explicit negative atoms $F^-$, and a set of rules $R$. A ground atom is a predicate applied to named entities, with no variables, for example $\texttt{has\_deployment\_approval}(\text{Service F})$. It also has a query atom $q=P(e)$, where $P$ is the predicate being asked about and $e$ is the entity. We write the predicate vocabulary as $\mathcal{V}$. Under \lcwa{}, the instance also gives a set $C \subseteq \mathcal{V}$ of complete predicates. The verifier first applies the rules in the positive direction. Starting from $F^+$, it repeatedly adds every rule conclusion whose conditions are already true. We write the resulting set as $\closure_R(F^+)$. This is the positive closure of the instance. The gold label for $q$ is then assigned in this order:
\begin{enumerate}
  \item If $q \in \closure_R(F^+)$, the label is \texttt{true}.
  \item Else, if $q \in F^-$, the label is \texttt{false}.
  \item Else, $q$ is missing. Missing queries are labeled according to the active closure contract:
  \begin{itemize}
    \item under \owa{}, the label is \texttt{unknown};
    \item under \cwa{}, the label is \texttt{false};
    \item under \lcwa{}, the label is \texttt{false} if $P \in C$, and \texttt{unknown} otherwise.
  \end{itemize}
\end{enumerate}
This procedure follows the semantics of definite logic programs. Rule bodies
contain no negation, so $\closure_R(F^+)$ is the least fixpoint of forward
chaining, and the \cwa{} label for a missing atom is the usual
negation-as-failure reading on the minimal model \citep{reiter1978closed}.
The one departure is $F^-$, which the verifier checks before any closure
applies. \exampleref{sec:example} shows this procedure on a concrete
instance.
We exclude target atoms that are both derivable and explicitly negated, since they would mix closure reasoning with the separate problem of resolving a conflict between a rule and an explicit fact, which is out of scope. The verifier does not reason backwards. If a rule condition is missing, the rule does not fire, and any \texttt{false} label for a missing query comes from the active closure contract, not a failed rule.

\subsection{Benchmark Design}
\label{sec:benchmark}

\bench{} has 960 English instances built from 320 base scenarios. Each scenario appears under all three closure contracts. All three variants use the same facts, rules, explicit negatives, and target atom. The scenarios use eight domains, with 120 instances in each one. These domains are {\em cloud deployment}, {\em finance controls}, {\em hospital access}, {\em library services}, {\em university registration}, {\em procurement review}, {\em manufacturing quality}, and {\em robotics operations}. 
%We chose them because a missing entry can mean unknown or false, depending on whether the source is complete. 
The instances mix missing direct facts and missing rule antecedents with distractors and controls where the target is entailed or explicitly negated. The base dataset has a {\em core contrastive set} and two {\em control groups}: of the 320 base scenarios, 192 are core contrastive scenarios, where changing the closure contract can change the gold label, and 128 are control scenarios whose gold label is stable across all three contracts (64 with a target that is entailed and 64 with a target that is explicitly negated). Each instance stores a gold answer and a reason type, so the answer can be traced to a fact derived by a rule, an explicit negative, or a missing target read using the active closure contract. We built the dataset from domain templates. For each domain, templates define the ontology and the case families used to generate the scenarios. A fixed script expanded these templates into base scenarios and created the three semantic variants for each scenario. We reviewed the generated blocks and ran a validation script that checks structural consistency, semantic labels, predicate completeness, and prompt contract agreement. When grouped by scenario, control labels stay the same under all three contracts, while core contrastive labels change only because the closure contract changes. The symbolic case stays fixed, but the active closure contract changes how the missing target is read. The extensions keep this base structure while moving the closure contrast into three agent formats. An example instance is discussed in \exampleref{sec:example} and reproduced in full in \appendixref{apd:example}.

\subsection{Agent Extensions}
\label{sec:extensions}

The base benchmark asks for a truth value. We add three agent extensions
where the same closure contract decides something else. In \emph{Ask/Act} it
decides an action, in \emph{SourceScope} a choice between sources of
different completeness, and in \emph{ClosureShift} an answer revised across
turns. The labels still come from the same symbolic truth labels and closure
contracts used in the base dataset.

\paragraph{Ask/Act.}

The \bench{} {\em Ask/Act} extension keeps the base instance and adds a decision. True targets become approve decisions, false targets become deny decisions, and unknown targets become requests for information. The symbolic case and truth label stay fixed. For request information instances, the dataset records whether the model should ask about the target itself or the first missing rule condition. The model also returns the missing field when needed, a question, and a rationale.

\paragraph{SourceScope.}

The \bench{} {\em SourceScope} extension splits evidence between two sources. Source A is complete only for the listed predicates. Source B is partial, so a missing fact from Source B does not support a false answer by itself. The coordinator must answer the query and justify both source choice and absence treatment. This extension has 360 instances from 120 base scenarios, and each selected scenario keeps the same \owa{}/\cwa{}/\lcwa{} triple.

\paragraph{ClosureShift.}

The \bench{} \emph{ClosureShift} extension tests whether the model updates its answer when completeness changes. Each dialogue has three turns. The facts, rules, glossary, and target stay fixed, while the complete predicates change. Turn 1 leaves every predicate open, so missing facts stay unknown. Turn 2 makes the full vocabulary complete, so missing facts count as false. Turn 3 replaces that full set with a smaller local set. In half of the dialogues, the local set still covers the target, so the answer should stay false. In the other half, the local set no longer covers the target, so the answer should return to unknown. The model returns the current answer, whether it changed, the update type, and a rationale.

\begin{example}
\label{sec:example}
For example, one dataset instance says that Service F passed unit tests. It also says that a service may deploy to production when it passed unit tests and has deployment approval. The target asks whether Service F may deploy to production. The antecedent \texttt{has\_deployment\_approval} and the target \texttt{may\_deploy} are different predicates. The instance never says whether Service F has deployment approval, so the rule cannot fire and the target is not derived. The three variants keep the same fact, rule, and target, and only the closure contract changes. The same base instance drives the three extensions. In \emph{Ask/Act} the \owa{} variant becomes an ask while \cwa{} and \lcwa{} become a deny. In \emph{ClosureShift} the answer shifts from unknown to false and back as the complete set changes across turns. In \emph{SourceScope} absence counts against a complete source but not an open one.
\end{example}

\section{Empirical Evaluation}
\label{sec:results-discussion}
This section presents the experimental setup and the results on the benchmark and its agent extensions.

\subsection{Experimental Setup}
\label{sec:experimental-setup}

We evaluate six model configurations, namely DeepSeek V4 Flash and DeepSeek V4 Pro \citep{deepseek2026v4}, Mistral Small 2603 \citep{mistral2026small4}, Mistral Medium 3.5 \citep{mistral2026medium35}, Llama 4 Scout \citep{meta2025llama4scout}, and Llama 4 Maverick \citep{meta2025llama4maverick}\footnote{In the tables, we shorten these names to DeepSeek Flash, DeepSeek Pro, Mistral Small, Mistral Medium, Llama Scout, and Llama Maverick.}. The runners call OpenAI compatible chat completion APIs at temperature 0. The system prompt restricts the model to the symbolic specification and closure contract and requires a single JSON object with an \texttt{answer} field and a \texttt{rationale}. The answer must be \texttt{true}, \texttt{false}, or \texttt{unknown}. We read the \texttt{answer} field and score any unparseable instance as incorrect. All results are mean $\pm$ sample standard deviation over three full runs. Hosted APIs are not fully deterministic even at temperature 0, because of effects such as nondeterministic kernels and server side batching. The three runs therefore measure how stable a model is from one run to the next, not decoding noise, and the deviations are small. We also rerun the base benchmark at temperature 1 as a robustness probe. All reported base and Ask/Act runs score their full 960 instance datasets. ClosureShift runs score all 100 dialogues and 300 turns. The 100 dialogues are sampled, with a fixed seed, from the core contrastive base scenarios, balanced evenly between dialogues that persist after narrowing and dialogues that reopen. SourceScope runs score all 360 extension instances, built from 120 base scenarios. The base dataset also stores train, development, and test splits by base scenario for future model development. The comparisons here use the full suite, and \sectionref{sec:limitations} explains why we report these numbers rather than results on the held out test split.

\subsection{Base Benchmark Results and Robustness}
\label{sec:base-results}

Control instances can make overall accuracy look higher when the closure contract does not affect the answer, and accuracy on single instances can hide whether a model got all variants of one base scenario right; behavioral testing makes the same point \citep{ribeiro2020checklist}. The base table therefore reports closure switch accuracy, which counts a base scenario correct only when all three variants are correct (\appendixref{apd:metrics}), and core switch restricts this to the 192 core contrastive scenarios. The two \lcwa{} scope metrics test whether the model says \texttt{false} only where local closure applies. We also compute plain \owa{} and \cwa{} accuracy, which range from $87.19\%$ to $100.00\%$ and from $69.27\%$ to $99.69\%$ across runs, but omit them from the main table since the switch and \lcwa{} metrics test the closure question more directly. The two fixed baselines reach $70.00\%$ and $90.00\%$ overall accuracy, respectively, but both obtain $0\%$ core switch accuracy because neither predicts all three closure variants correctly for any core scenario. In \tableref{tab:base-results}, overall accuracy makes several models look stronger than they are on closure. DeepSeek Pro reaches $93.61\%$ overall accuracy, but only $77.43\%$ core switch accuracy. Mistral Medium has the highest overall accuracy after Llama Maverick, but its \lcwa{} closed scope accuracy is $56.95\%$. In these cases, it often answers \texttt{unknown} even when the prompt says the relevant predicate is complete and absence should count as false. DeepSeek Flash, DeepSeek Pro, and Mistral Medium get all \lcwa{} open scope cases right in the base runs, but they miss many closed scope cases. They leave the answer as \texttt{unknown} when local closure should make it false. Mistral Small and Llama Scout are less steady. They fail to close complete predicates and sometimes close predicates that should stay open. Llama Maverick is a useful positive control because it nearly solves the suite. This makes it less likely that the instances are unclear or mislabeled. As a robustness check, we rerun the base benchmark at temperature 1 for three full passes per model. Temperature 1 does not remove the closure errors. Overall accuracy moves by less than one point for four models, which matches earlier reports that temperature barely affects aggregate accuracy on problem solving \citep{renze2024temperature}, while the closure sensitive metrics are less stable, with DeepSeek Flash and DeepSeek Pro losing the most core switch accuracy. We therefore use temperature 0 for the main comparison, \appendixref{apd:temperature} reports the full temperature 1 comparison. 

\begin{table}[t]
\floatconts
  {tab:base-results}
  {\caption{Base \bench{} results over three runs with temperature 0.}}
  {\small
  \begin{tabular*}{\linewidth}{@{\extracolsep{\fill}}lccccc@{}}
  \toprule
  Model & Overall & Closure switch & Core switch & \lcwa{} closed & \lcwa{} open \\
  \midrule
  Mistral Small & $80.38 \pm 1.02$ & $55.73 \pm 2.43$ & $27.60 \pm 3.65$ & $31.60 \pm 4.34$ & $31.25 \pm 1.80$ \\
  DeepSeek Flash & $90.31 \pm 0.47$ & $81.25 \pm 0.83$ & $68.75 \pm 1.38$ & $49.65 \pm 3.18$ & $100.00 \pm 0.00$ \\
  DeepSeek Pro & $93.61 \pm 0.21$ & $86.46 \pm 0.18$ & $77.43 \pm 0.30$ & $62.15 \pm 1.20$ & $100.00 \pm 0.00$ \\
  Mistral Medium & $95.59 \pm 0.37$ & $87.08 \pm 0.65$ & $78.47 \pm 1.09$ & $56.95 \pm 2.17$ & $100.00 \pm 0.00$ \\
  Llama Scout & $71.18 \pm 0.53$ & $50.73 \pm 0.48$ & $18.23 \pm 1.04$ & $4.51 \pm 1.59$ & $52.43 \pm 3.01$ \\
  Llama Maverick & $99.86 \pm 0.06$ & $99.59 \pm 0.18$ & $99.31 \pm 0.30$ & $99.65 \pm 0.60$ & $100.00 \pm 0.00$ \\
  \bottomrule
  \end{tabular*}}
\end{table}

\subsection{Agent Extension Results}
\label{sec:agent-results}
The same missing information errors also appear in the agent extensions. \appendixref{apd:metrics} defines every column.

\paragraph{Ask/Act.}
As defined in \sectionref{sec:extensions}, \emph{Ask/Act} maps truth labels to actions. Its columns, action accuracy, core switch action, Ask F1, and excessive ask, are defined in \appendixref{apd:metrics}. Excessive ask is the fraction of closure false cases, where the gold action is \texttt{deny}, on which the model instead requests more information. \tableref{tab:ask-act} shows that truth errors become action errors. In closure false cases, the complete source already rules out the target, so the model should deny. When it treats the answer as unknown, it asks for more information even though the source is already enough. Excessive ask is above $20\%$ for DeepSeek Flash, DeepSeek Pro, and Mistral Medium, and above $45\%$ for Llama Scout. Robot planners built on language models show the same failure, where the agent asks for help even though the evidence already settles the decision \citep{ren2023robots}. A model that reads absence under a complete predicate as \texttt{unknown} requests information even though it should deny.

\begin{table}[t]
\floatconts
  {tab:ask-act}
  {\caption{Ask/Act results over three runs with temperature 0. Excessive ask measures closure false cases where the gold action is \texttt{deny}, but the model requests more information.}}
  {\small
  \begin{tabular*}{\linewidth}{@{\extracolsep{\fill}}lcccc@{}}
  \toprule
  Model & Action acc. & Core switch action & Ask F1 & Excessive ask \\
  \midrule
  Mistral Small & $83.44 \pm 0.55$ & $32.81 \pm 1.80$ & $76.06 \pm 1.01$ & $39.00 \pm 0.72$ \\
  DeepSeek Flash & $93.09 \pm 0.16$ & $77.43 \pm 0.30$ & $89.49 \pm 0.22$ & $21.07 \pm 0.53$ \\
  DeepSeek Pro & $93.47 \pm 0.16$ & $80.73 \pm 0.52$ & $90.19 \pm 0.21$ & $21.76 \pm 0.53$ \\
  Mistral Medium & $92.98 \pm 0.16$ & $72.57 \pm 0.79$ & $89.53 \pm 0.21$ & $23.38 \pm 0.53$ \\
  Llama Scout & $71.63 \pm 0.75$ & $12.33 \pm 1.59$ & $55.65 \pm 0.61$ & $45.60 \pm 2.01$ \\
  Llama Maverick & $100.00 \pm 0.00$ & $100.00 \pm 0.00$ & $100.00 \pm 0.00$ & $0.00 \pm 0.00$ \\
  \bottomrule
  \end{tabular*}}
\end{table}

\paragraph{SourceScope.}
In \emph{SourceScope} full accuracy requires the truth value, the source, and the treatment of absence all to be correct. Leakage means treating absence from an open source as closed, and Loss means treating absence from a complete source as open. \tableref{tab:sourcescope} shows that a model can get the truth value right but still choose the wrong source or handle absence in the wrong way. Mistral Medium reaches $91.58 \pm 0.16\%$ truth accuracy, but only $79.26 \pm 5.61\%$ full accuracy. DeepSeek Flash has low closure loss, $4.72 \pm 0.00\%$, but high leakage, $29.70 \pm 2.93\%$. It sometimes treats absence from an open source as false. DeepSeek Pro has no leakage in the three runs, but still loses closure for absence from a complete source at $14.47 \pm 1.44\%$. Leakage and loss are opposite source scope errors. The model either closes a source that should remain open or fails to close a source declared complete. This helps explain why truth accuracy can remain high while full accuracy drops.

\begin{table}[ht]
\floatconts
  {tab:sourcescope}
  {\caption{SourceScope results over three runs with temperature 0. Lower is better for Leakage and Loss.}}
  {\small
  \setlength{\tabcolsep}{2pt}
  \begin{tabular*}{\linewidth}{@{\extracolsep{\fill}}lcccccc@{}}
  \toprule
  Model & Full & Truth & Source & Absence & Leakage & Loss \\
  \midrule
  Mistral Small & $61.76 \pm 6.72$ & $76.20 \pm 4.02$ & $68.24 \pm 3.77$ & $82.69 \pm 3.22$ & $8.48 \pm 2.78$ & $77.04 \pm 11.18$ \\
  DeepSeek Flash & $85.19 \pm 4.65$ & $89.54 \pm 0.89$ & $97.78 \pm 0.28$ & $92.96 \pm 1.79$ & $29.70 \pm 2.93$ & $4.72 \pm 0.00$ \\
  DeepSeek Pro & $93.52 \pm 2.43$ & $95.74 \pm 0.42$ & $96.39 \pm 2.17$ & $95.93 \pm 1.67$ & $0.00 \pm 0.00$ & $14.47 \pm 1.44$ \\
  Mistral Medium & $79.26 \pm 5.61$ & $91.58 \pm 0.16$ & $93.15 \pm 0.58$ & $84.82 \pm 4.89$ & $0.00 \pm 0.00$ & $28.62 \pm 0.55$ \\
  Llama Scout & $70.65 \pm 1.67$ & $76.85 \pm 0.42$ & $77.87 \pm 1.67$ & $74.35 \pm 0.58$ & $2.73 \pm 1.82$ & $71.07 \pm 1.09$ \\
  Llama Maverick & $95.37 \pm 0.16$ & $98.98 \pm 0.89$ & $97.13 \pm 0.42$ & $97.13 \pm 0.42$ & $0.00 \pm 0.00$ & $3.46 \pm 3.03$ \\
  \bottomrule
  \end{tabular*}}
\end{table}

\paragraph{ClosureShift.}
In \emph{ClosureShift}, dialogue accuracy requires all three turn answers correct, while revision and persistence split the update turns by whether the gold answer should change or stay the same. \tableref{tab:closureshift} shows that most models handle the move from open to closed, with almost perfect revision accuracy. The harder case is persistence after narrowing. This turns the open/closed predicate distinction into an update problem, instead of a static query-answering problem \citep{ahmetaj2024cqa,marconi2025openclosed,oudshoorn2026shacl}. When the global complete set is replaced by a local set that still contains the relevant predicate, several models drop the false answer and return to \texttt{unknown}, the retention failure in the last column. The main failure is not reopening after closure disappears, but rather losing
\texttt{false} when local closure still applies. \figureref{fig:closure-error-profile} summarizes the same closure failure across the base benchmark and the three agent extensions.

\begin{table}[ht]
\floatconts
  {tab:closureshift}
  {\caption{ClosureShift results over three runs with temperature 0. Lower is better for Retention failure.}}
  {\small
  \begin{tabular*}{\linewidth}{@{\extracolsep{\fill}}lcccc@{}}
  \toprule
  Model & Dialogue acc. & Revision & Persistence & Retention failure \\
  \midrule
  Mistral Small & $58.00 \pm 2.65$ & $100.00 \pm 0.00$ & $21.33 \pm 4.16$ & $78.67 \pm 4.16$ \\
  DeepSeek Flash & $91.00 \pm 1.00$ & $100.00 \pm 0.00$ & $82.00 \pm 2.00$ & $18.00 \pm 2.00$ \\
  DeepSeek Pro & $98.00 \pm 1.00$ & $100.00 \pm 0.00$ & $96.00 \pm 2.00$ & $4.00 \pm 2.00$ \\
  Mistral Medium & $83.33 \pm 0.58$ & $100.00 \pm 0.00$ & $66.67 \pm 1.15$ & $33.33 \pm 1.15$ \\
  Llama Scout & $50.00 \pm 0.00$ & $99.33 \pm 0.67$ & $2.00 \pm 2.00$ & $97.33 \pm 3.06$ \\
  Llama Maverick & $100.00 \pm 0.00$ & $100.00 \pm 0.00$ & $100.00 \pm 0.00$ & $0.00 \pm 0.00$ \\
  \bottomrule
  \end{tabular*}}
\end{table}

\subsection{Limitations}
\label{sec:limitations}
\bench{} is template generated, which makes its labels auditable but restricts linguistic diversity. We do not test naturalistic paraphrases or a parser--verifier neurosymbolic baseline. The results therefore measure adherence to closure contracts rather than general real-world reasoning. Likewise, the agent extensions have limitations. They extend the closure problem to action choice, source scope, and updates across turns, but they do not execute tools nor do they engage with a real world environment. They can thus best be thought of as controlled experiments designed to understand agent behavior. They cannot take the place of testing deployed agent systems in practice. For our study, the full suite is necessary due to the focus on identifying a failure mode. However, for model selection in the future, it may be better to use the held-out test split. In any future study, once this benchmark becomes publicly available, care should be taken to check for training leakage. 

% Divergent heat-map palette from image
\definecolor{errDarkBlue}{HTML}{2066A8}
\definecolor{errMedBlue}{HTML}{8EC1DA}
\definecolor{errLightBlue}{HTML}{CDE1EC}
\definecolor{errGray}{HTML}{EDEDED}
\definecolor{errLightRed}{HTML}{F6D6C2}
\definecolor{errMedRed}{HTML}{D47264}
\definecolor{errDarkRed}{HTML}{AE282C}

% Lower is better: blue = low error, red = high error
\newcommand{\errA}[1]{\cellcolor{errDarkBlue!85}{#1}}
\newcommand{\errB}[1]{\cellcolor{errMedBlue!85}{#1}}
\newcommand{\errC}[1]{\cellcolor{errLightBlue!85}{#1}}
\newcommand{\errD}[1]{\cellcolor{errLightRed!85}{#1}}
\newcommand{\errE}[1]{\cellcolor{errMedRed!85}{#1}}
\newcommand{\errF}[1]{\cellcolor{errDarkRed!85}{#1}}

\begin{figure}[ht]
\floatconts
{fig:closure-error-profile}
{\caption{Closure error profile across \bench{} extensions. Colors use a diverging palette, blue for low error and red for
high error \citep{simplifiedscience2024palettes}.}}
{\small
\setlength{\tabcolsep}{3pt}
\renewcommand{\arraystretch}{1.18}
\begin{tabular*}{\linewidth}{@{\extracolsep{\fill}}lccccc@{}}
\toprule
Model &
\shortstack{Base\\\lcwa{} closed\\error} &
\shortstack{Ask/Act\\excessive\\ask} &
\shortstack{SourceScope\\loss} &
\shortstack{SourceScope\\leakage} &
\shortstack{ClosureShift\\retention\\failure} \\
\midrule
Mistral Small   & \errE{68.40} & \errC{39.00} & \errE{77.04} & \errB{8.48}  & \errE{78.67} \\
DeepSeek Flash  & \errD{50.35} & \errC{21.07} & \errA{4.72}  & \errC{29.70} & \errB{18.00} \\
DeepSeek Pro    & \errC{37.85} & \errC{21.76} & \errB{14.47} & \errA{0.00}  & \errA{4.00} \\
Mistral Medium  & \errD{43.05} & \errC{23.38} & \errC{28.62} & \errA{0.00}  & \errC{33.33} \\
Llama Scout     & \errF{95.49} & \errD{45.60} & \errE{71.07} & \errA{2.73}  & \errF{97.33} \\
Llama Maverick  & \errA{0.35}  & \errA{0.00}  & \errA{3.46}  & \errA{0.00}  & \errA{0.00} \\
\bottomrule
\end{tabular*}
}
\end{figure}
\section{Conclusion}
\label{sec:conclusion}
\bench{} tests this setting directly by asking whether an LLM can follow a stated symbolic closure assumption when it answers or acts from incomplete records. To do this, the benchmark fixes the symbolic case and varies only the interpretation of absence required by the closure contract. Because control instances keep the same label under every contract, a model can score high
overall accuracy while ignoring the closure contract entirely. A baseline that always reads absence under one fixed contract reaches $70.00\%$ overall accuracy with $0\%$ core switch accuracy. Multiple models respond correctly to control questions while making mistakes when the closure contract changes the meaning of absence. Such failures appear particularly clearly under \lcwa{}, where closure should be applied only to the complete predicates. Under \lcwa{}, models tend to give an \texttt{unknown} response even though the prompt declares the relevant predicate complete, so the correct answer is \texttt{false}. In \emph{Ask/Act}, the mistake can change the decision: the model requests additional information when it should deny. In \emph{SourceScope}, the model may still return the correct truth value while using the wrong source or while applying closure where it does not belong. This is a problem because agents, tools, and knowledge bases may have different closure assumptions. A coordinator in a multi-agent system needs to know when absence from a complete source supports denial and when it should remain \texttt{unknown}. In \emph{ClosureShift}, we observe a related issue. Models usually handle the change from open to closed, yet several lose \texttt{false} as the set of complete predicates shrinks while still covering the target. These results show that evaluations of agent-like models should test not only reasoning over given facts but also the handling of missing information. The remaining
errors point to a narrower weakness: models can use stated facts and rules but still fail to apply the closure contract that determines what absence means.

\newpage

\bibliography{references}

\appendix

\section{Use of generative AI}
\label{apd:genai}
AI tools were used only for spelling, grammar, wording, and consistency checks while preparing this manuscript. They were not used to create benchmark labels, score model outputs, choose which results to report, or perform the analysis.

%\section{Dataset and code}
%\label{apd:repro}

%The anonymized repository is available during review and will be released publicly upon publication:

%{\small
%\url{https://anonymous.4open.science/r/ClosureBench-nesyai-1492/}
%}

\section{Validation checks}
\label{apd:validation}
The dataset validator checks structural consistency, semantic labels, predicate completeness constraints, and prompt contract agreement. The validation report records 960 instances (called \emph{items} in the released data and scripts), 320 base scenarios, 320 instances per closure contract, and validation status PASS.

\section{Metric definitions}
\label{apd:metrics}
Every metric is computed by the symbolic scoring scripts and reported as a percentage.

\paragraph{Base benchmark.}
\emph{\lcwa{} closed scope} and \emph{\lcwa{} open scope} are instance accuracy on the \lcwa{} instances where the queried predicate is complete, so absence should be \texttt{false}, and where it is open, so absence should be \texttt{unknown}. \emph{Instance consistency} is the fraction of instances with the same answer across the three runs.

\paragraph{Ask/Act.}
\emph{Action accuracy} is the fraction of instances whose predicted action (\texttt{approve}, \texttt{deny}, or \texttt{request\_information}) matches the gold action, and \emph{core switch action} applies the closure switch metric to this action match over the core contrastive scenarios. \emph{Ask F1} is the F1 score (precision and recall) of \texttt{request\_information} predictions against the gold requests. \emph{Excessive ask} is the fraction of closure false instances, whose gold action is \texttt{deny}, on which the model instead returns \texttt{request\_information}.

\paragraph{SourceScope.}
Each instance has a gold truth value, a gold source, and a gold absence label (\texttt{open\_absence} or \texttt{closed\_absence}). \emph{Truth}, \emph{Source}, and \emph{Absence} are the accuracy of each field, and \emph{Full} requires all three correct on one instance. \emph{Leakage} is the rate on \texttt{open\_absence} instances at which the model closes a source that is still open, and \emph{Loss} is the rate on \texttt{closed\_absence} instances at which it fails to close a complete source.

\paragraph{ClosureShift.}
Each dialogue has three turns. \emph{Dialogue accuracy} is the fraction of dialogues correct on all three turns. \emph{Revision} and \emph{Persistence} are instance accuracy on the update turns whose gold answer should change and should stay the same. \emph{Reopening accuracy} is accuracy on the turn 3 dialogues where the narrowed set no longer covers the target. \emph{Retention failure} is the rate, on turn 3 dialogues that persist after narrowing with gold \texttt{false}, at which the model returns \texttt{unknown}.

\section{Complete worked example}
\label{apd:example}

This appendix reproduces the base instance of \exampleref{sec:example} exactly as sent to the model. In the released dataset it is base scenario \texttt{cloud\_\allowbreak deployment\_\allowbreak 0257\_\allowbreak closed\_\allowbreak derived\_\allowbreak missing\_\allowbreak antecedent}, shown under its \owa{}, \cwa{}, and \lcwa{} variants. The three variants share the same facts, rule, glossary, and target, and only the \emph{Closure contract} block and \emph{Complete predicates} line differ, so we give the \owa{} prompt in full and only the changed lines for \cwa{} and \lcwa{}. Long lines are wrapped only for display.

\paragraph{\owa{} variant, gold answer \texttt{unknown}.}
{\small
\begin{verbatim}
You are evaluating a formal knowledge-base statement.
Return exactly one JSON object with keys answer and rationale.
The answer must be one of: true, false, unknown.

Formal rule semantics:
- Positive facts are true.
- Negative facts make the corresponding target atom false.
- A rule derives its conclusion only when every antecedent is true.
- First apply all rules until no new facts can be derived.
- Then apply the semantic contract to unstated and underivable atoms.

Semantic contract: Open-world assumption.
Predicates are open by default. A target atom that is not explicitly
stated, explicitly negated, or derivable from the rules is unknown.
Complete predicates for this item: none.

Important closure rule:
For a complete predicate P, if P(entity) is not explicitly stated and
cannot be derived after applying the rules, then P(entity) is false.
This does not require an explicit negative fact.

Predicate glossary:
- has_deployment_approval: has deployment approval
- has_incident_note: has an incident note
- may_deploy: may deploy to production
- needs_sre_review: needs SRE review
- passed_unit_tests: passed unit tests

Positive facts:
- Service F passed unit tests.

Negative facts:
- None

Rules:
- If a service passed unit tests and has deployment approval, then
  that service may deploy to production.

Target atom: may_deploy::Service F.
Target statement: Service F may deploy to production.
\end{verbatim}
}

\paragraph{\cwa{} variant, gold answer \texttt{false}.}
{\small
\begin{verbatim}
Semantic contract: Closed-world assumption.
All predicates in the vocabulary are complete. After applying the
rules, every unstated and underivable atom is false.
Complete predicates for this item: has_deployment_approval,
has_incident_note, may_deploy, needs_sre_review, passed_unit_tests.
\end{verbatim}
}

\paragraph{\lcwa{} variant, gold answer \texttt{false}.}
{\small
\begin{verbatim}
Semantic contract: Locally closed-world assumption.
Only the predicates listed as complete are closed. After applying the
rules, an unstated and underivable atom whose predicate is complete is
false; an unstated and underivable atom whose predicate is not
complete is unknown.
Complete predicates for this item: has_deployment_approval, may_deploy.
\end{verbatim}
}

\paragraph{Expected outputs and a typical error.}
The rule cannot fire because its antecedent (deployment approval for Service F) is missing, so the target is never derived. The correct \texttt{answer} is \texttt{unknown} under \owa{} and \texttt{false} under \cwa{} and \lcwa{}, since \texttt{may\_deploy} is complete in both. The common error is to answer \texttt{unknown} on the \lcwa{} variant, where the model leaves the target open even though \texttt{may\_deploy} is closed and one of its antecedents is absent. This is what drives the low \lcwa{} closed scope and core switch numbers in \tableref{tab:base-results}.

\section{Temperature 1 robustness}
\label{apd:temperature}

To check whether deterministic decoding masks stochastic fragility, we rerun the 960 instance base benchmark at temperature 1 for three full passes per model. \tableref{tab:temperature-one} compares the temperature 1 means with the temperature 0 means from \tableref{tab:base-results}. Temperature 1 does not remove the closure errors. Overall accuracy changes by less than one point for four models, which matches earlier reports that temperature barely affects aggregate accuracy on problem solving \citep{renze2024temperature}. The closure sensitive metrics move more, so sampling exposes instability that overall accuracy does not. DeepSeek Flash loses $9.55$ points of core switch accuracy, and DeepSeek Pro loses $16.32$ points. Mistral Medium and Llama Maverick remain comparatively stable, while Mistral Small gains on \lcwa{} closed scope but remains far below the strong models. Several models still look accurate while failing the closure switch cases, so temperature 1 shows which models are less stable without changing the main result.

\begin{table}[t]
\floatconts
  {tab:temperature-one}
  {\caption{Temperature robustness comparison for the base benchmark}}
  {\small
  \begin{tabular*}{\linewidth}{@{\extracolsep{\fill}}lcccccc@{}}
  \toprule
  Model & \multicolumn{2}{c}{Overall} & \multicolumn{2}{c}{Core switch} & \multicolumn{2}{c}{\lcwa{} closed} \\
  \cmidrule(lr){2-3}\cmidrule(lr){4-5}\cmidrule(l){6-7}
   & T0 & T1 & T0 & T1 & T0 & T1 \\
  \midrule
  Mistral Small & $80.38$ & $80.59$ & $27.60$ & $29.17$ & $31.60$ & $40.62$ \\
  DeepSeek Flash & $90.31$ & $88.71$ & $68.75$ & $59.20$ & $49.65$ & $51.39$ \\
  DeepSeek Pro & $93.61$ & $90.55$ & $77.43$ & $61.11$ & $62.15$ & $56.60$ \\
  Mistral Medium & $95.59$ & $95.31$ & $78.47$ & $77.60$ & $56.95$ & $56.94$ \\
  Llama Scout & $71.18$ & $71.04$ & $18.23$ & $18.58$ & $4.51$ & $3.47$ \\
  Llama Maverick & $99.86$ & $99.55$ & $99.31$ & $97.92$ & $99.65$ & $99.31$ \\
  \bottomrule
  \end{tabular*}}
\end{table}

\end{document}